# Prospect Theory in Large Language Models: A Meta-Analysis of Loss Aversion, Diminishing Sensitivity, and Probability Weighting

Adi Tarabeih

Technion – Israel Institute of Technology, Israel, Haifa

*Address for correspondence: Adi Tarabeih, Faculty of Data and Decision Sciences,

Technion - Israel Institute of Technology, Haifa 3200003, Israel.

Email: adi.tarabeih@campus.technion.ac.il

**Prospect Theory in Large Language Models: A Meta-Analysis of Loss Aversion, Diminishing Sensitivity, and Probability Weighting**

## Abstract

Prospect theory splits risky choice into three parts: loss aversion, measured by the relative weight of losses to gains, λ (e.g., λ = 2 implies that losses weigh twice as much as gains), diminishing sensitivity to the size of outcomes (i.e., the curvature and linearity of the value function), and the nonlinear weighting of probabilities. I examined whether Large Language Models (LLMs) display the same properties as people by meta-analyzing 102 fitted estimates from 11 studies covering 32 language models. The three components emerge differently. Loss aversion is modest: the pooled λ is 1.26 (95% CI 1.06 to 1.46), far below the strongest human meta-analytic estimate (λ = 1.96) and compatible with more conservative ones (1.07 and 1.31). Matched by design, LLMs are compatible with people when gains and losses are symmetric (λ = 1.15 against 1.07), but well below them when losses are smaller than gains (1.46 against 2.3), where human loss aversion is strongest. Value functions are close to linear rather than concave under the Tversky–Kahneman method (1.01, against a human value of 0.67), but more concave than in people under the Tanaka method (0.52, against 0.67), so diminishing sensitivity depends on how it is measured. Probability weighting shows a human-like inverse S shape (γ = 0.81 under Tversky–Kahneman and 0.84 under the Tanaka method), though on few studies. In sum, LLMs show little loss aversion and little diminishing sensitivity, but retain human-like probability distortion, yet these generalizations shift when looking at specific model families and fitting methods.

## Introduction

Large language models (LLMs) have become an integral part of human decision making. Their role is twofold. On one side, LLMs are increasingly deployed to make or advise on consequential choices: screening job applicants and drafting hiring recommendations, assessing creditworthiness in lending, supporting triage and treatment choices in clinical settings, and allocating attention in financial and operational contexts (Eigner & Händler, 2024). On the other side, the same models are being used as virtual human participants, standing in for people in surveys, experiments, and pilot studies on the assumption that their responses approximate how humans would choose (Argyle et al., 2023; Aher, Arriaga & Kalai, 2023; Dillion, Tandon, Gu & Gray, 2023; Horton, 2023). Both uses rest on one unexamined premise: that a model's choices resemble a person's. Whether they actually do is an empirical question, and the present paper addresses this question in the context of decisions under risk, namely given known probabilities and outcomes for making choices.

The dominant account of human choice under risk is prospect theory (Kahneman & Tversky, 1979; Tversky & Kahneman, 1992), formalized around three systematic departures from expected value rationality. The first is loss aversion, the tendency for losses to weigh more heavily than equivalent gains. The second is diminishing sensitivity, a concave value function in which each additional unit of outcome matters less as magnitudes grow. The third is nonlinear probability weighting, where small probabilities are given more weight than they objectively deserve, producing a weighting curve shaped like an inverse S. This overweighting of rare events is well documented when the probabilities are stated explicitly, that is, in decisions from description; when the same choices are instead learned through sampling, in decisions from experience, the pattern reverses and rare events tend to be underweighted, a discrepancy known as the description-experience gap (Hertwig, Barron, Weber & Erev, 2004; Hertwig & Erev, 2009). Because almost all of the LLM literature uses described gambles, the relevant human comparison is the description case.

In the human literature, diminishing sensitivity and nonlinear probability weighting remain uncontested. Studies robustly show a concave value function and an inverse S probability weighting curve. Yet there is still some ambiguity regarding the magnitude of the value curvature and of the weighting distortion across studies and populations (Ruggeri et al., 2020). For loss aversion the findings in humans are far more disparate: three recent meta-analyses place the human coefficient anywhere from roughly 1.07 (Yechiam & Zeif, 2025), through 1.31 (Walasek, Mullett & Stewart, 2024), to 1.96 (Brown, Imai, Vieider & Camerer, 2024), a range wide enough to fuel an active debate concerning the existence and magnitude of loss aversion in humans.

Whether LLMs display these same signatures matters precisely because of their two roles. If models have absorbed human biases, then those biases now operate at scale inside real decisions about hiring, credit, and care, and they do so invisibly. If models do not share these biases, then using them as proxies for human participants is unfounded, and results drawn from virtual samples will not generalize to people. Either way, the size and even the sign of the gap between model and human behavior is something to be measured rather than assumed.

Two hypotheses bracket what one might expect. The first is continuity. Because LLMs are trained on enormous amounts of text produced by people, they are in effect trained to reproduce human judgment (see e.g., McShane, Gal, and Duhachek, 2025) and should therefore exhibit the same prospect theory distortions at comparable magnitudes. The second is divergence. Because these models are also optimized to be helpful and accurate, and because the newest reasoning models are trained to deliberate step by step rather than answer reflexively, they may instead move toward normative, expected value rationality and show weaker distortions than people, or none at all (Chen, Liu, Shan & Zhong, 2023). These two accounts make opposite predictions about the most recent, most capable models.

Both hypotheses also treat language models as a single kind of agent, which they are not. Models from different developers, such as OpenAI's GPT, Anthropic's Claude, Google's Gemini and Alibaba's Qwen, are trained on different data, aligned to human preferences with different fine tuning procedures, and built at different scales, and these choices shape what the models say and choose (Ouyang et al., 2022; Santurkar et al., 2023). Studies that compare many models on the same tasks find large differences, for example models answered classic reasoning tasks inconsistently, and differently from one another as well as from people (Macmillan-Scott & Musolesi, 2024), and across 35 models, open source models took fewer risks than proprietary ones (Coda-Forno et al., 2024). Even within one family, successive generations can behave very differently, as when the intuitive reasoning errors of earlier GPT models disappeared in ChatGPT (Hagendorff, Fabi & Kosinski, 2023). The studies reviewed here show the same pattern for risky choice: given identical gambles, GPT, Claude and Gemini models produced different degrees of loss aversion. An average across models may therefore hide large differences between families, so I treat model family, together with release year, as a moderator in its own right.

Adjudicating between these two accounts, and asking whether the answer differs from one family of models to another, requires bringing the scattered individual estimates into a single common frame. That is the aim of the present work. I conducted a systematic review and a random effects meta-analysis of every prospect theory parameter that has been fitted to a language model's own risky monetary choices, pooling 102 fitted estimates from 11 studies spanning 32 models, separately for loss aversion, diminishing sensitivity, and probability weighting. To foreshadow the results, the picture that emerges is not uniform. Loss aversion is modest: the pooled coefficient sits at the low end of the human range, far below the strongest human estimate, and compared with people tested in the same designs, the models are compatible with them when gains and losses are symmetric but fall well short of them when losses are smaller than gains. Diminishing sensitivity depends on how it is measured: value functions are close to linear under one fitting method and more concave than in people under the other. Probability weighting is the one component that tracks human behavior, with the inverse S shape found in people, although the evidence for it rests on few studies. And the differences between models are so large that the average describes few of them. In other words, the answer to whether LLMs think like humans about risk is neither yes nor no; it differs by component and by model.

## Method

### *Data collection*

My data collection and reporting follows the PRISMA 2020 reporting guidance (Page et al., 2021). The functional forms of the value and probability-weighting functions, and the meaning of each parameter (λ, α, β, σ, γ, φ), are defined in Appendix A. A study was included in the meta-analysis only if it reported at least one of three possible fitted prospect-theory parameters: loss aversion (λ), diminishing sensitivity (α, β or σ), or probability weighting (γ or φ).

Records were identified through the OpenAlex works endpoint (api.openalex.org), which indexes journal articles, conference proceedings, arXiv and SSRN. In addition, Google Scholar and Semantic Scholar served as supplementary cross-checks on coverage[1]. The search was run on 3 September 2026 and used the following key words at the title and abstract level: ("large language model" OR "large language models" OR "LLM" OR "LLMs" OR "ChatGPT" OR "GPT-4" OR "GPT-3.5" OR "GPT-4o" OR "language model" OR "generative AI" OR "foundation model") AND ("prospect theory" OR "cumulative prospect theory" OR "loss aversion" OR "certainty equivalent" OR "probability weighting" OR "risky choice" OR "decision under risk" OR "risk preference" OR "lottery" OR "gamble"). The search was restricted to 2021–2026, a window that excludes no eligible work because LLMs capable of these tasks were not publicly available beforehand. Three further studies were added outside the database search: Yechiam and Tarabeih (2026), a study I coauthored; the Master's thesis of Waisman (2026), obtained directly from Eldad Yechiam; and Yee and Koh (2026), identified by citation searching.

Each record was screened using the following inclusion criteria: First and foremost, at the study level, whether the report estimates at least one prospect-theory parameter for a language model's own risky monetary choices. I therefore did not include studies reporting only accuracy, pass rates, choice proportions or qualitative bias labels without a fitted parameter. Secondly, with respect to measures I only used (a) monetary risky-choice problems with stated outcomes and probabilities, (b) Parameters obtained by maximum likelihood, least squares or an equivalent structural fitting procedure, with the parameterization method stated in the report, and (c) I removed estimates whose fitted parameter simply sat at the edge of the range the fitting routine would allow (using the specific bounds reported in each paper). These could well be non-converged fits rather than genuine measurements, so I excluded them from the pooled estimates and used them only in a robustness check. Wang, R. et al. (2025) themselves describe two of their boundary λ values (0.01) as unusual and as indicating irregular fits.

[1] Because Google Scholar matches full text rather than title and abstract, the same term pairs there return far larger pools dominated by incidental mentions (1,150 records for "large language model" with "prospect theory"; 2,830 with "lottery"), and neither cross-check surfaced a qualifying study absent from the OpenAlex set.

Finally, at the population level, I focused on LLMs, generative or encoder-based, of any scale or vendor, while excluding non-language-model agents (e.g., classical RL, symbolic solvers) unless an LLM arm was also reported. Titles and abstracts were first screened with the assistance of a large language model (Claude, Anthropic), which was given the inclusion criteria above and recorded a decision and a reason for every record; next, I verified the included set at full text. To check the model's screening, I independently screened a random 20% sample of the records (80 records, drawn from a rerun of the same search), blind to the model's decisions. The model and I agreed on 67 of the 80 records (84%; Cohen's $\kappa = 0.49$). I then read the 13 records on which we disagreed in full: twelve did not meet the inclusion criteria, and one was an SSRN version of a study already included through citation searching (Yee & Koh, 2026), so no eligible study was lost.

From each of the included studies, I recorded the available parameters and the sample size needed for approximating their sampling distribution (i.e., number of agent runs). I also recorded the task used to elicit choices (certainty-equivalent, direct choice, or the Tanaka multiple-pricelist)2 and whether each

[2] The three task types differ only in how the model is asked to reveal its preferences. In the certainty equivalent task the model chooses, one at a time, between a gamble and a run of fixed sure amounts; the point where it gives up the gamble for the sure amount shows what the gamble is worth to it, and the value and weighting curves are fitted to those switch points (for example, a 30% chance of $40 and otherwise $10, weighed against rising sure amounts). In direct choice the model simply accepts or rejects a single bet, with the possible loss raised step by step, and loss aversion is read off as the largest loss it will still accept against a fixed win (for example, “a coin flip that wins $6 or

estimate used the Tversky–Kahneman or the Tanaka fitting method. Both are shown in Table 1 but were not entered as moderators. Because the two fitting methods put the parameters on different scales (Appendix A), I pooled curvature and probability weighting separately for each method (loss aversion is a single ratio and is pooled across methods). I then coded a set of moderators that might explain differences between studies, in two classes. The first class was Model-identity moderators which include (A) the model's family (GPT, Claude, Gemini, Qwen, and smaller open-source families) and (B) its release year. The second class of moderators was Study-design moderators which included (i) the stake size, that is, the largest monetary outcome in the gambles a study used (from \$10 to \$5,000; Wang, Z. et al., 2025, used points, counted at face value as 150), entered on a log scale because the stakes span almost three orders of magnitude, and relevant because in people loss aversion is stronger when more is at stake (Ert & Erev, 2013); (ii) the gain–loss structure of its stimuli, that is, whether the losses were smaller than, equal to, or larger than the gains (L < G / L = G / L > G). This matters because designs with smaller losses than gains, though common in the literature, can inflate measured loss aversion through contrast effects (Ert & Erev, 2013; Zeif & Yechiam, 2022), whereas symmetric gains and losses avoid that confound; (iii) whether the model was persona or personality-conditioned rather than answering natively. Persona or personality conditioning refers to studies that, instead of querying the model in its default state, first instruct it to adopt a specific identity or personality before it answers: for example to role-play a given age or gender (Ross et al., 2024; Corazzini et al., 2025), a described investor persona (Liu et al., 2025), or a Big-Five personality profile (Hartley et al., 2025). I coded whether each reported estimate came from such a conditioned prompt or from the model's native, unconditioned responses. In the meta-regressions, these moderators were entered in a fixed order, in five nested models: Model 1 includes only model family, Model 2 adds release year, Model 3 adds the gain–loss structure, Model 4 adds stake size, and Model 5 adds persona conditioning, so comparing successive models shows how much each block explains beyond the previous ones. I entered loses \$5: accept or reject?"). In the Tanaka multiple price list the model works down three sets of paired lotteries that gradually tilt toward the riskier option, and the rows where it switches fix curvature, probability weighting and loss aversion at once.

the properties of the model first, because whether prospect theory distortions belong to the model is the central question, and added the study design moderators afterwards to see how much of the remaining heterogeneity reflects how the studies were designed. Among these, the gain–loss structure and stake size came first because they describe the gambles themselves, whereas persona conditioning changes who is answering them. The gain–loss structure and stake size are mainly properties of a study's gambles rather than of a model, so their effects rest mostly on differences between studies. The gain–loss structure was coded as losses smaller than gains ($L < G$) against equal losses and gains ($L = G$). It applies only to loss aversion, because it requires gambles that mix gains and losses, so Model 3 is not estimated for diminishing sensitivity and probability weighting.

**Human comparison values**

I compared each parameter against meta-analyzed published human values. For loss aversion these were considerably disparate in different meta-analyses, running from $\lambda = 1.96$ (Brown, Imai, Vieider & Camerer, 2024), $\lambda = 1.31$ (Walasek, Mullett & Stewart, 2024), and down to $\lambda = 1.07$ (Yechiam & Zeif, 2025). Differences likely stemmed from the disparate inclusion criteria: Brown et al. (2024) summarized an eclectic set of 150 studies, while Walasek et al. focused on a smaller corpus of studies that included the complete data, and Yechiam and Zeif summarized a subgroup of studies with symmetric and unordered gains and losses (which they argue avoids potential confounds). Yechiam and Zeif's (2025) re-analysis further shows the human coefficient is design-conditional, near 2.3 for ordered designs with losses smaller than gains and near 1.07 for symmetric, unordered designs. To my knowledge it is the only source of human values broken down by design, so the comparisons by design below rest on it. I examined how LLMs stand in comparison to these three benchmarks of human loss aversion, and also compared the estimates from each gain–loss structure with the human value for that structure. For diminishing sensitivity and probability weighting the main benchmark is the human meta-analysis of Imai et al. (2025), which gives a value curvature of about $\alpha = 0.67$ and a probability weighting of $\gamma = 0.68$. Imai et al. (2025) is my primary benchmark for curvature and weighting; I also report the original estimates of Tversky and Kahneman (1992), $\alpha = 0.88$ and $\gamma = 0.65$ (Table S1). For the Tanaka parameterization the benchmark is the human sample of Tanaka, Camerer and Nguyen (2010), $\sigma = 0.67$ and $\gamma = 0.685$; in their parameterization $\sigma$ is the exponent of the power value function, so a lower $\sigma$ means a more concave function. Where a benchmark was published with a confidence interval (Brown et al., 2024: 1.82 to 2.11; Yechiam & Zeif, 2025: 0.97 to 1.18; Imai et al., 2025, for $\gamma$: 0.66 to 0.70), the test of the difference includes its uncertainty; the other benchmarks were treated as fixed values. Because a nonsignificant difference does not establish equivalence, I describe such cases as compatible with the benchmark rather than equal to it.

**Estimation Method**

For each parameter, I first present the fitted estimates descriptively and then pool them with a random effects model, using inverse variance weights (Hedges & Olkin, 2014). The between estimate variance $\tau^2$ was estimated by restricted maximum likelihood (REML) rather than with the moment estimator of DerSimonian and Laird (1986), and all tests used the adjustment of Knapp and Hartung (2003), which replaces the normal distribution with a t distribution and widens the confidence intervals when the number of estimates is small. Besides the confidence interval of each pooled estimate, I report the 95% prediction interval, the range in which the estimate for a new model is expected to fall; when heterogeneity is as large as here, the prediction interval describes the dispersion across models better than the pooled mean does. The model was fitted separately to each parameter and, for curvature and weighting, separately within each parameterization, because the Tversky–Kahneman and Tanaka scales differ. Each pooled estimate was tested against the no-distortion value of 1 and against each human benchmark, and heterogeneity was summarized with $\tau$, Cochran's Q and $I^2$ (Higgins & Thompson, 2002). When a study reported a confidence interval or the standard deviation across runs for a parameter (Hartley et al., 2025; Jia et al., 2024; Wang, R. et al., 2025; Ge et al., 2026), I used it; otherwise I imputed a standard error with a rule adapted from Furukawa et al. (2006): the standard deviation of the estimates in the pooled set divided by the square root of the number of runs behind the estimate. Because this standard deviation reflects differences between models rather than sampling error within a study, and because repeated runs of one model are not independent participants, the imputed standard errors are small relative to $\tau$. The random effects weights are therefore nearly equal across estimates, which is why the equal weight analyses in Table S2 give almost the same results, and the very high $I^2$ values partly reflect this construction rather than a precise measure of heterogeneity. Like the other two parameters, loss aversion was pooled and analysed on its original scale, so every $\lambda$ reported in the text, tables and figures is an ordinary $\lambda$ value. For loss aversion the unit of analysis is the individual estimate ($k = 28$), because stake size and persona conditioning are properties of an estimate rather than of a model: the same model was often tested under different designs in different studies. Pooling the 19 models instead is reported as a robustness check.

Because the estimates are nested (several studies reported more than one model, and several models were tested by more than one study), I report three units of analysis. The individual estimate is the main unit. In the model level analysis, the estimates of a model tested by more than one study were first combined into a fixed effect average, and the models were then pooled with the same random effects model. In the study level analysis, each study contributes one effect (the mean of its estimates, with the standard error of that mean) and the studies are pooled in the same way; because publication and small study effects act on studies, this is the level at which I assess publication bias. To account directly for the dependence between estimates from the same study, I also fitted a multilevel model in which estimates are nested in studies (REML), with cluster robust standard errors and a small sample correction (the factor $m/(m - p)$ and t tests with $m - p$ degrees of freedom, where m is the number of studies and p the number of coefficients; Viechtbauer, 2010). With ten studies for loss aversion, and six and four for curvature and weighting, this model is reported as a robustness check; for curvature and weighting the cluster robust tests could not be computed, because the number of studies is not larger than the number of coefficients.

To examine factors that account for the heterogeneity, I ran meta-regressions with the same REML estimator and the Knapp and Hartung adjustment, adding the moderators one at a time in a fixed order across nested models (Models 1 to 5); for each I report the coefficient, an omnibus F test, and $R^2$ (the share of $\tau^2$ explained). In every meta-regression, GPT is the reference category for model family, and release year and stake size are centered, so each coefficient describes a difference from GPT models of average release year tested at average stakes. Loss aversion is a single ratio that is comparable across fitting methods, so its meta-regression includes all studies; its coefficients are differences in $\lambda$. For diminishing sensitivity and probability weighting the meta-regressions include only the Tversky–Kahneman studies, because the Tanaka subset (two studies) is too small to support a regression. The gain–loss structure entered the loss aversion regression as Model 3; in addition, I compared the models that Waisman (2026) ran under both structures, so that only the structure varied. Because these regressions treat the estimates as independent, I repeated the loss aversion regressions with the multilevel model, and I checked the stability of the main coefficients by leaving out one study at a time, by comparing families within the studies that tested more than one family (study fixed effects), and by excluding RoBERTa, the only encoder model in the corpus (Supplementary Table S3). Analyses used Python (NumPy and SciPy). Because I implemented the estimators myself, I checked them against independent implementations: the DerSimonian–Laird estimates match those of the statsmodels package, and the REML estimates of $\tau^2$ match those obtained by direct numerical maximization of the restricted likelihood, to four decimals.

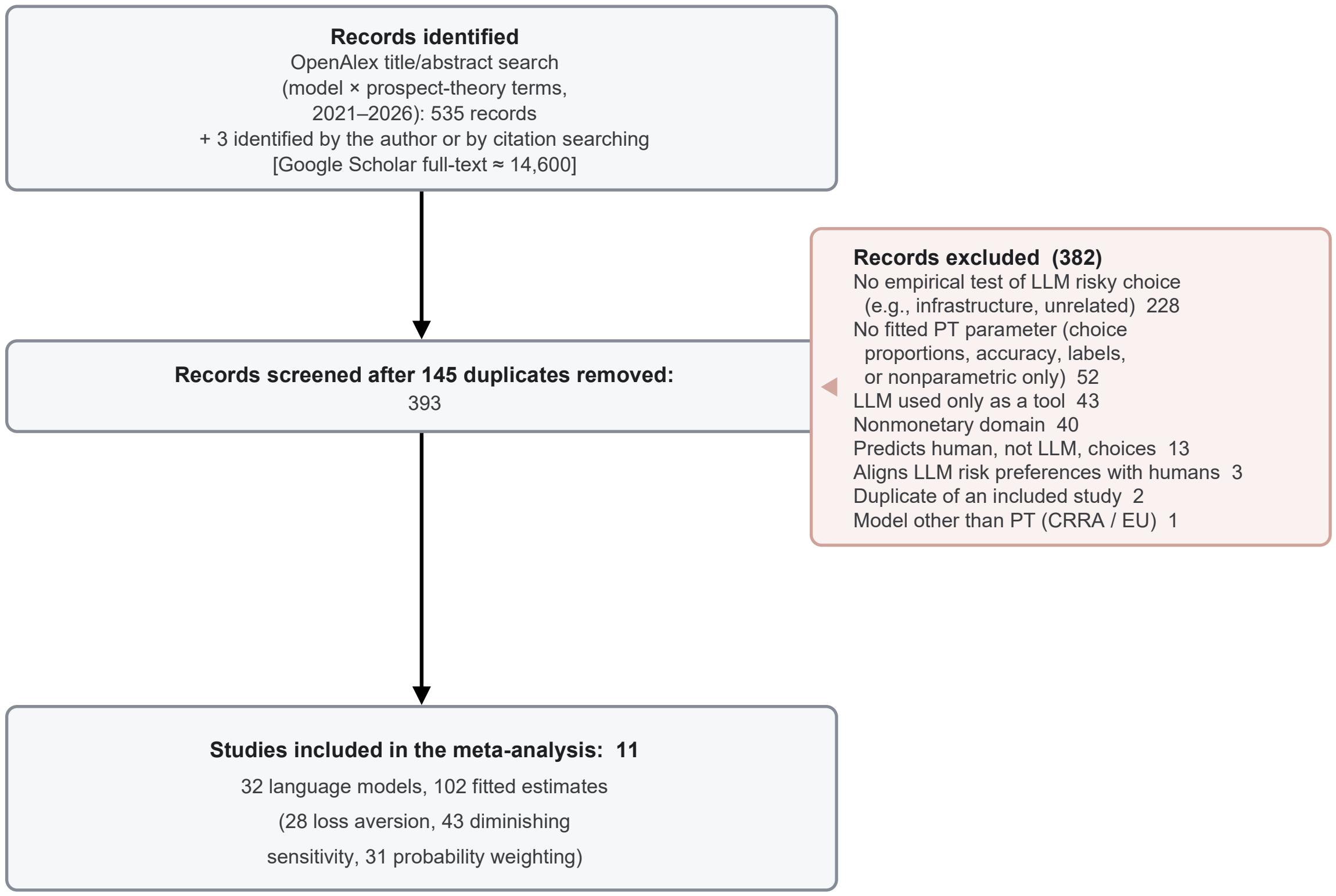

**Figure 1.** PRISMA flow diagram of study selection.

### The included Studies

The search yielded 102 fitted estimates from 11 studies covering 32 language models. In sum 28 estimates were computed for loss aversion, 43 for diminishing sensitivity and 31 for probability weighting (Figure 1). The models came from the GPT, Claude, Gemini, Llama, Qwen, and Mistral families, plus the encoder-based RoBERTa. Regarding task type, four studies used certainty-equivalent, five used direct choice, and two used the Tanaka multiple-price-list (Table 1). With respect to gain–loss structure, most loss-aversion studies used symmetric gains and losses (L = G), whereas the two Tanaka multiple-price-list studies and the sequential gamble study of Wang, Z. et al. (2025) used mostly smaller losses than gains (L < G), and Waisman (2026) ran the same models under both structures. Six extreme estimates – four for loss aversion and two for weighting – running to the edge of the fitting routine were excluded, and used only in the robustness test reported below.

Eleven studies with fitted prospect-theory parameters remained. Eight of the eleven fitted a diminishing-sensitivity parameter (six under the Tversky–Kahneman method and two under Tanaka, whose weighting curve differs), six of them also fitting a probability-weighting parameter, while the remaining three measured only loss aversion.

**Table 1. Studies included in the quantitative synthesis**

| Study | Models | Task type | Parameter | N (Agents) |
|---|---|---|---|---|
| Ross, Kim, & Lo (2024) | GPT-4, GPT-4 Turbo | Certainty-equivalent | λ, α, β | 100 |
| Jia et al. (2024) | GPT-4-Turbo, Claude-3-Opus, Gemini-1.0-Pro (Claude-3-Opus λ: boundary fit) | Multiple-price-list (Tanaka) | λ, σ, γ | 300 |
| Hartley et al. (2025) | GPT-4o, GPT-4o mini, Claude 3 Sonnet, Claude 3 Haiku | Certainty-equivalent | λ, α, γ | 15 |
| Corazzini et al. (2025) | Qwen-2.5-32B-Instruct | Certainty-equivalent | λ, α, β, γ | 100 |
| Wang, R. et al. (2025) | Llama-3.1-8B, Mistral-7B, Qwen2.5-7B/14B/32B, GPT-5-Mini, Gemini-2.5-Flash | Multiple-price-list (Tanaka) | σ, λ, γ | 256 |
| Wang, Z. et al. (2025) | GPT-3.5, GPT-4.1 | Direct choice | λ | 30 |
| Liu et al. (2025) | Llama3-8B, OLMo-2-7B | Certainty-equivalent | λ, α, β | 400 |
| Ge, Zhang, & Vorobeychik (2026) | Qwen2.5-7B/Math-7B, Qwen3-30B-Instruct/Thinking, Olmo-3-7B-Instruct/Think, Olmo-3-32B-Think, Gemini-2.5-Pro/Flash, DeepSeek-Chat/R1, GPT-5.1, GPT-4.1, Claude-4.5-Haiku | Direct choice | α (their σ), γ | 10 |
| Yee & Koh (2026) | GPT-4o, GPT-4o mini, Gemini-2.5-Pro, Claude-3.5-Haiku | Direct choice | λ | 100 |
| Waisman (2026) | GPT-4o, GPT-3.5, RoBERTa | Direct choice | λ, α, β, γ | 100 |
| Yechiam & Tarabeih (2026) | GPT-4o, Qwen2.5-72B | Direct choice | λ | 200 |

## Results

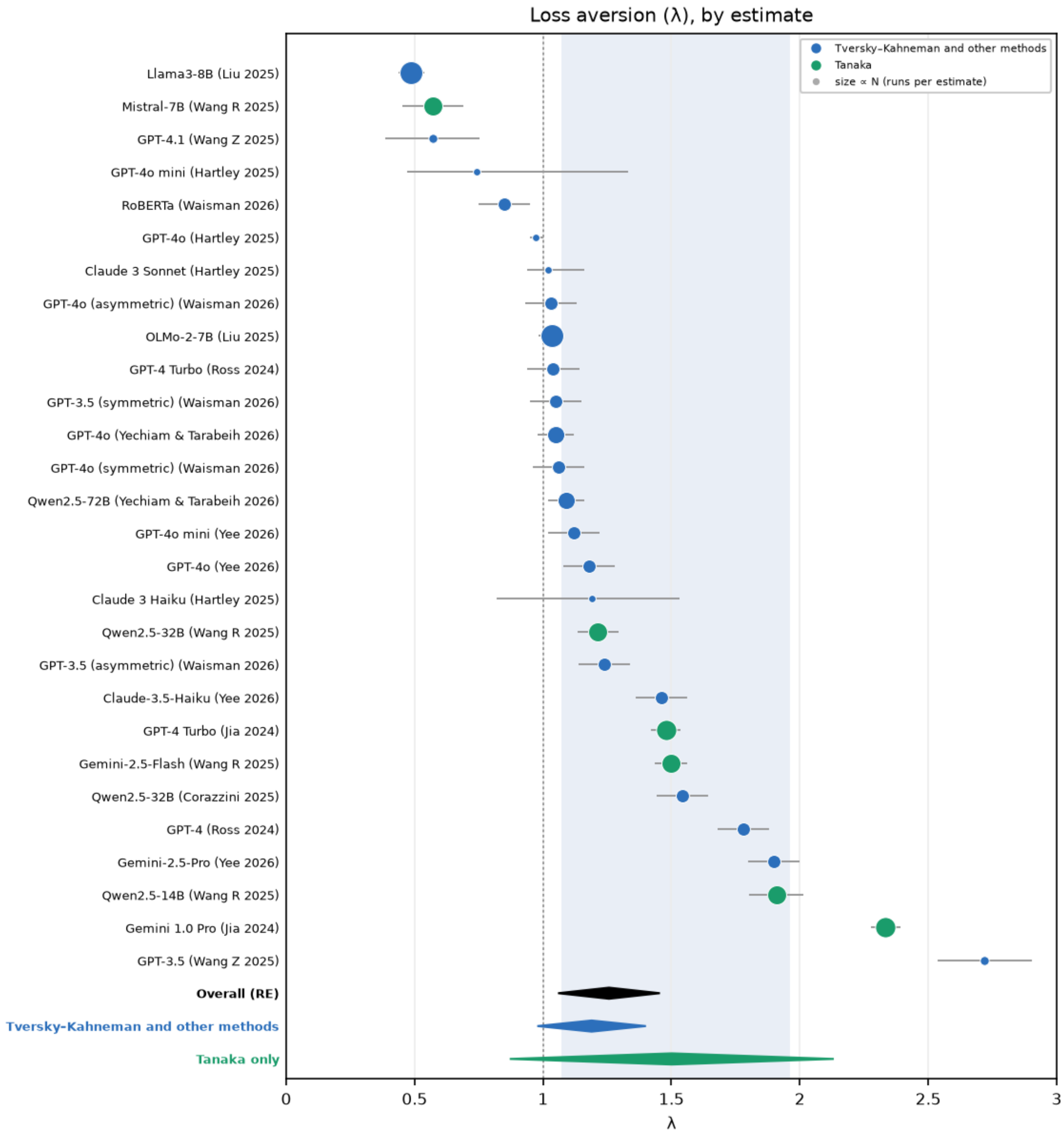

**Figure 2.** Loss aversion forest plot by model and study. Values are λ on its original scale. The shaded band marks the range of human estimates (1.07 to 1.96).

**Loss aversion**

Figure 2 presents the loss-aversion forest plot, showing every model's estimate and the pooled value. Each row is one estimate, labelled by model and study: the dot is the estimate (blue for the Tversky–Kahneman and other methods, green for Tanaka; dot size is proportional to the number of runs behind the estimate) and the horizontal line its 95% confidence interval. The dotted vertical line marks $\lambda = 1$ (no loss aversion), the shaded band the range of human estimates, and the diamonds at the bottom the pooled estimates. The 28 estimates differ widely, from 0.49 (Llama3-8B-Instruct) to 2.72 (GPT-3.5)14 show significant loss aversion, 5 the reverse, and 9 no significant asymmetry between gains and losses. The overall loss aversion coefficient was $\lambda = 1.26$, 95% CI [1.06, 1.46], a small but significant degree of loss aversion (against 1: $t(27) = 2.66$, $p = .013$; Figure 2). It falls well below the human value of Brown et al. (2024), 1.96 ($t = -5.78$, $p < .001$, taking the uncertainty of both estimates into account), and is compatible with the 1.31 of Walasek et al. (2024; $t = -0.55$, $p = .59$) and with the 1.07 of Yechiam and Zeif (2025; $t = 1.69$, $p = .10$). The estimates were very heterogeneous ($\tau = 0.51$; the Q and $I^2$ values in Table S1 partly reflect the imputed standard errors, see Method), and the 95% prediction interval runs from 0.20 to 2.31: a new model could plausibly be loss seeking, loss neutral, or as loss averse as the strongest human estimate. Pooling the 19 models instead of the 28 estimates gives the same value, $\lambda = 1.26$ [1.02, 1.49].

To see what explains this heterogeneity, I entered the moderators in five nested meta-regressions (Table 2). Model 1 contains only model family; Model 2 adds release year, Model 3 adds the gain–loss structure, Model 4 adds stake size and Model 5 adds persona conditioning, so comparing the models shows what each block explains beyond the previous ones. Model family accounted for 27% of the variance between estimates ($R^2 = .27$; $F(4, 23) = 3.29$, $p = .029$): $\lambda$ was about 0.7 higher in Gemini models than in GPT models (Model 1: +0.69, $p = .021$), and about 0.5 lower in the four small open source models (Llama, Mistral, OLMo and RoBERTa; −0.49, $p = .063$), whereas Claude and Qwen did not differ from GPT. Holding family constant, $\lambda$ fell by about 0.16 for every year of release (Model 2: $p = .032$; $R^2 = .38$). The gain–loss structure (Model 3) added nothing on its own (L < G: +0.16, $p = .35$). Stake size (Model 4) was significant once family and release year were held constant: $\lambda$ rose by about 0.35 for every tenfold increase in the stakes ($p = .030$), in the same direction as in people, but stake size alone did not predict loss aversion ($p = .79$). Without Wang, Z. et al. (2025), whose stakes were in points, the stake coefficient fell to +0.23 ($p = .08$). With stakes held constant, the designs with smaller losses than gains gave a somewhat higher $\lambda$ (+0.27 to +0.30), but the difference was not significant ($p = .08$ in Model 5). Persona conditioning (Model 5) was not significant.

These regressions treat the 28 estimates as independent, although they come from ten studies. I therefore repeated them with a multilevel model in which estimates are nested in studies, with cluster robust standard errors, and checked them by leaving out one study at a time, by comparing families within studies, and by excluding RoBERTa (Supplementary Table S3). Model family remained significant and the pooled $\lambda$ hardly changed. The coefficients were stable throughout, including the decline with release year, but the cluster robust tests rest on ten studies and few degrees of freedom, so they are not powerful; the Gemini difference reached significance in only some of them, and the release year decline lost significance when Wang, Z. et al. (2025) were excluded. I therefore regard the higher loss aversion of Gemini models as suggestive, the decline with release year as resting substantially on the GPT lineage of Wang, Z. et al. (2025), and the effects of the study design moderators as exploratory.

Because human loss aversion itself depends on the design (see Human comparison values), I also compared the estimates from each gain–loss structure with the human value for that structure (Waisman, 2026, contributes to both). In the 18 estimates from seven studies in which gains and losses were symmetric ($L = G$), the pooled $\lambda$ was 1.15 [0.98, 1.32], compatible with the human value of 1.07 for such designs (Yechiam & Zeif, 2025; $t = 0.79$, $p = .44$) and well below that of Brown et al. (2024; $t = -7.39$, $p < .001$). In the ten estimates from four studies in which losses were smaller than gains ($L < G$), the pooled $\lambda$ was 1.46 [0.96, 1.96], well below the value of about 2.3 that people show in such designs ($t = -3.81$, $p = .004$). This difference would remain significant even if the human value of 2.3 carried a standard error of up to 0.30, that is, a 95% interval of roughly 1.7 to 2.9. The symmetric result also holds within each task: $\lambda$ = 1.10 [0.79, 1.40] in the certainty equivalent studies and 1.20 [0.96, 1.43] in the direct choice studies, in which the possible loss is raised step by step, both compatible with 1.07 ($p = .86$ and $p = .31$). The models are therefore compatible with people where human loss aversion is weakest, but do not show the rise in loss aversion that designs with smaller losses than gains produce in people, in line with the small and nonsignificant coefficient of the gain–loss structure in Table 2.

**Table 2. Loss aversion – nested meta-regression (all estimates; coefficients are differences in λ).**

| | **Model 1** | **Model 2** | **Model 3** | **Model 4** | **Model 5** |
|---|---|---|---|---|---|
| Intercept | 1.22 (0.12)*** | 1.20 (0.11)*** | 1.14 (0.13)*** | 1.10 (0.11)*** | 1.08 (0.11)*** |
| Claude | 0.00 (0.29) | 0.10 (0.27) | 0.15 (0.27) | 0.11 (0.25) | 0.15 (0.25) |
| Gemini | 0.69 (0.28)* | 0.84 (0.26)** | 0.78 (0.27)** | 0.85 (0.25)** | 0.86 (0.24)** |
| Qwen | 0.22 (0.25) | 0.31 (0.23) | 0.29 (0.23) | 0.35 (0.21) | 0.35 (0.21) |
| Other families | −0.49 (0.25) | −0.59 (0.23)* | −0.57 (0.23)* | −0.63 (0.21)** | −0.89 (0.31)** |
| Release year | | −0.16 (0.07)* | −0.16 (0.07)* | −0.22 (0.07)** | −0.25 (0.07)** |
| Losses smaller than gains (L < G) | | | 0.16 (0.17) | 0.27 (0.16) | 0.30 (0.16) |
| Stake size (log10) | | | | 0.35 (0.15)* | 0.31 (0.15) |
| Persona conditioning | | | | | 0.49 (0.42) |
| Model fit (F) | $F(4, 23) = 3.29$* | $F(5, 22) = 4.19$** | $F(6, 21) = 3.63$* | $F(7, 20) = 4.56$** | $F(8, 19) = 4.23$** |
| $R^2$ | 0.27 | 0.38 | 0.38 | 0.49 | 0.50 |
| *Estimates (k)* | 28 | 28 | 28 | 28 | 28 |

Notes: Each estimate is one unit (k = 28). Moderators added in order: Model 1: model family (reference GPT; the four single model families Llama, Mistral, OLMo and RoBERTa form one group), Model 2: release year (centred), Model 3: gain–loss structure (losses smaller than gains, L < G, against equal losses and gains), Model 4: stake size (log10 of the largest outcome in the gambles, centred; a coefficient is the change in λ for a tenfold increase in the stakes), and Model 5: persona conditioning. Coefficients are differences in λ, estimated with REML and the Knapp and Hartung adjustment, with SEs in parentheses; model fit is the omnibus F test. Supplementary Table S3 repeats the analysis with estimates nested in studies. * $p < .05$, ** $p < .01$, *** $p < .001$.

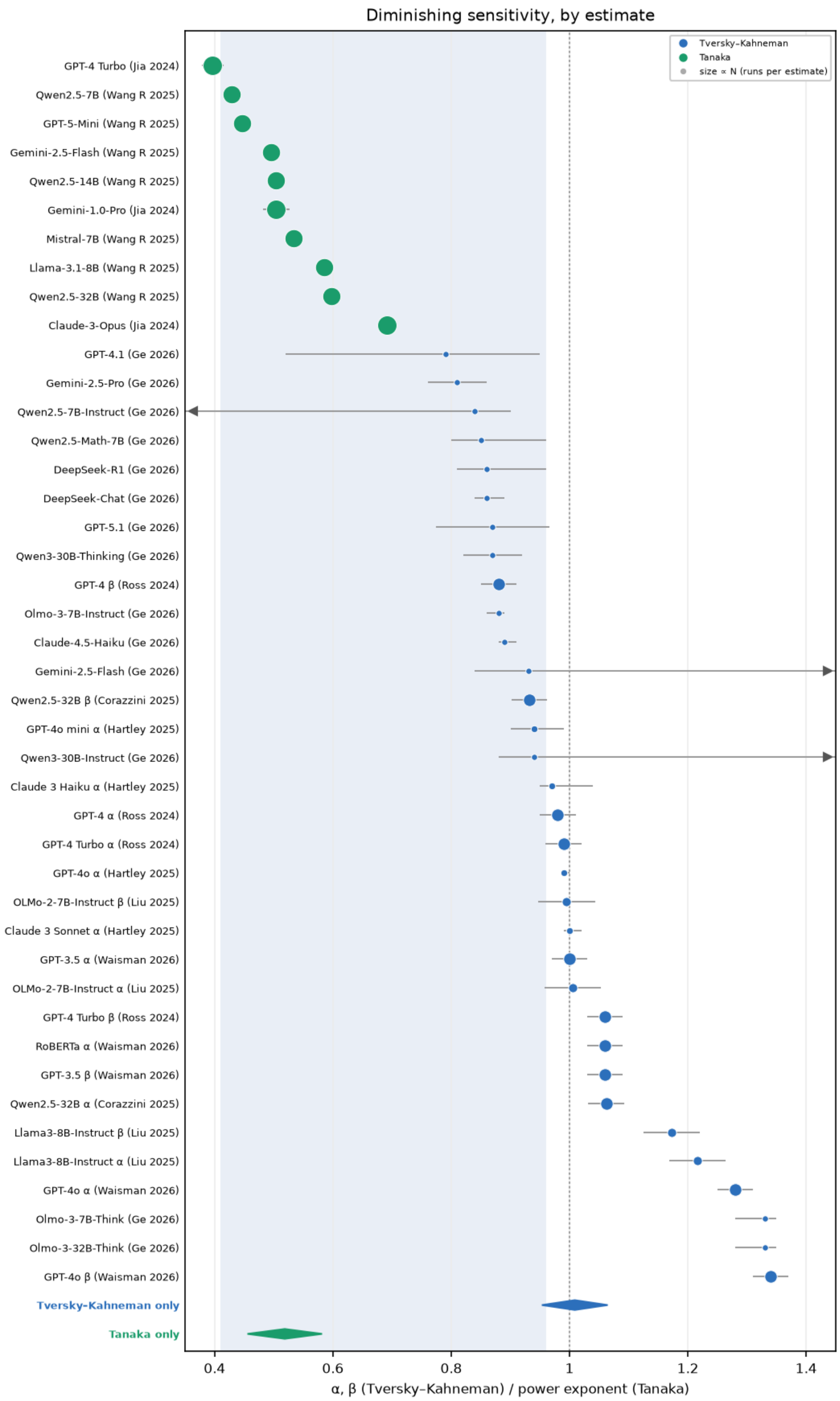

**Figure 3.** Forest plot - diminishing sensitivity . Tversky–Kahneman estimates (blue) are α, β or the curvature parameter of Ge et al. (2026); Tanaka estimates (green) are the exponent of the power value function. The shaded band marks the range of human estimates reported by Ruggeri et al. (2020), 0.41 to 0.96.

**Diminishing sensitivity**

Figure 3 presents the curvature forest plot, showing every model's estimate and the pooled value. Under the Tversky–Kahneman method the pooled curvature was 1.01 [0.95, 1.06], essentially a straight line (against 1: $t(32) = 0.32$, $p = .75$), and far less concave than the meta-analytic human value of 0.67 (Imai et al., 2025; $t = 12.41$, $p < .001$) and than the original estimate of Tversky and Kahneman (1992), 0.88 ($t = 4.72$, $p < .001$). Under this method, then, the models show little of the human concavity: they value larger outcomes roughly in proportion to their size. This holds for each component separately: the gain curvature $\alpha$ pooled to 1.04 [0.98, 1.11] ($k = 12$), the loss curvature $\beta$ to 1.06 [0.92, 1.21] ($k = 7$), and the curvature parameter of Ge et al. (2026) to 0.94 [0.83, 1.05] ($k = 14$). The prediction interval (0.69 to 1.33) shows that most models are close to linear. In the nested meta-regressions (Table 3), model family alone did not significantly account for the heterogeneity ($F(6, 26) = 2.12$, $p = .085$). The gain–loss structure (Model 3) could not be tested here: half of the curvature estimates come from gambles with only gains or only losses, where the structure is not defined, and all of the others come from designs with equal losses and gains. Estimates from larger stakes were more concave, as in people: the curvature fell by about 0.11 for every tenfold increase in the stakes (Model 4, $p = .005$), and once stake size was held constant newer models were slightly less concave (+0.06 per year of release, $p = .030$) and persona prompts slightly more concave (−0.23, $p = .014$, Model 5). Because stake size varies only between studies, and the largest stakes ($5,000) come from a single study (Ge et al., 2026), these effects largely contrast a few of the six studies; with so few studies they could not be tested with cluster robust standard errors, and I regard them as exploratory.

Under the Tanaka method, with all estimates expressed as the exponent of the power value function (Appendix A), the pooled curvature was 0.52 [0.46, 0.58] across ten estimates from two studies, more concave than the value of 0.67 that Tanaka, Camerer and Nguyen (2010) report for people ($t(9) = -5.43$, $p < .001$). The two methods thus place the models on opposite sides of the human benchmark: close to linear under Tversky–Kahneman, and more concave than people under Tanaka. Because the two methods were applied to different models, I cannot separate the fitting method from the sample; the safe conclusion is that curvature is not yet comparable across methods.

**Table 3. Diminishing sensitivity – nested meta-regression (α; Tversky–Kahneman studies only).**

| | Model 1 | Model 2 | Model 3 | Model 4 | Model 5 |
|---|---|---|---|---|---|
| Intercept | 1.02 (0.04)*** | 1.03 (0.04)*** | — | 0.99 (0.04)*** | 0.99 (0.04)*** |
| Qwen | -0.10 (0.08) | -0.11 (0.08) | — | -0.05 (0.07) | -0.04 (0.06) |
| OLMo | 0.08 (0.07) | 0.07 (0.08) | — | 0.12 (0.07) | 0.21 (0.07)** |
| Claude | -0.07 (0.09) | -0.08 (0.09) | — | -0.07 (0.08) | -0.09 (0.07) |
| DeepSeek | -0.16 (0.11) | -0.19 (0.11) | — | -0.09 (0.10) | -0.09 (0.09) |
| Gemini | -0.21 (0.14) | -0.23 (0.15) | — | -0.13 (0.13) | -0.13 (0.12) |
| Other | 0.13 (0.09) | 0.14 (0.09) | — | 0.17 (0.08)* | 0.35 (0.10)** |
| Release year | | 0.01 (0.03) | — | 0.06 (0.03)* | 0.09 (0.03)** |
| Stake size (log10) | | | — | -0.11 (0.03)** | -0.14 (0.03)*** |
| Persona conditioning | | | — | | -0.23 (0.09)* |
| Model fit (F) | F(6, 26) = 2.12 | F(7, 25) = 1.82 | — | F(8, 24) = 3.33* | F(9, 23) = 4.46** |
| $R^2$ | 0.17 | 0.15 | — | 0.35 | 0.48 |
| Estimates (k) | 33 | 33 | — | 33 | 33 |

Notes: Only studies fitted with the Tversky–Kahneman method are included, because the Tanaka curvature parameter (σ) is on a different scale from the Tversky–Kahneman parameter (α). Moderators added in order: Model 1: model family, Model 2: release year, Model 3: gain–loss structure, not estimated because these gambles do not mix gains and losses, Model 4: stake size (log10, centred), and Model 5: persona conditioning. Coefficients from REML with the Knapp and Hartung adjustment; model fit is the omnibus F test. With six studies the cluster robust tests could not be computed, so the table is exploratory. Other = Llama 3 and RoBERTa. SEs in parentheses.

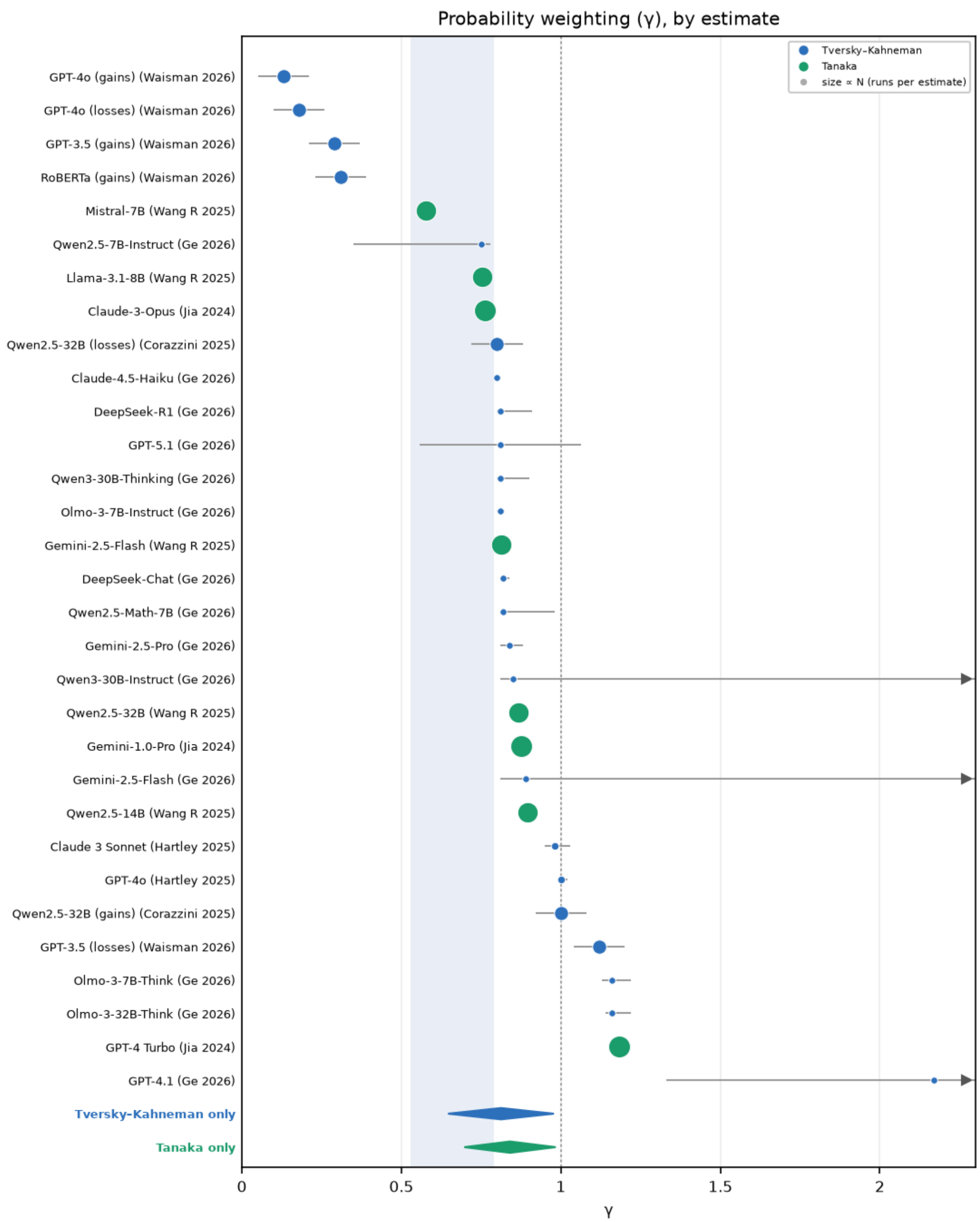

**Figure 4.** Forest plot for probability weighting . The shaded band marks the range of human estimates reported by Ruggeri et al. (2020), 0.53 to 0.79.

**Overweighting of small probabilities**

Figure 4 presents the weighting forest plot, showing every model's estimate and the pooled value. Under the Tversky–Kahneman method the pooled estimate was $\gamma = 0.81$ [0.65, 0.98], significantly below one ($t(22) = -2.35$, $p = .028$), the inverse S shape found in people, and compatible with the meta-analytic human value of Imai et al. (2025), 0.68 ($t = 1.64$, $p = .12$), and with the original estimate of Tversky and Kahneman (1992), 0.65 ($t = 2.03$, $p = .055$). Without the single outlying estimate of GPT-4.1 ($\gamma = 2.17$), the pool is 0.77 [0.64, 0.91]. Under the Tanaka method the pooled estimate was $\gamma = 0.84$ [0.70, 0.98] across eight estimates from two studies, also below one ($t(7) = -2.63$, $p = .034$) and slightly above the human value of Tanaka et al. (2010), 0.685 ($t = 2.57$, $p = .037$); adding back the two boundary fits of Wang, R. et al. (2025) moves the pool to 1.32 [0.57, 2.07] (see Publication bias). Heterogeneity was very high ($\tau = 0.35$), and the prediction interval under Tversky–Kahneman (0.06 to 1.56) includes both strong distortion and none. In the nested meta-regressions (Table 4; Model 3 could not be tested, for the same reason as for curvature, and Model 5 because no weighting estimate came from a persona prompt), no model was significant (all omnibus $p > .47$). Newer reasoning models did not weight probabilities differently from conversational models: GPT-5.1, DeepSeek-R1 and Qwen3-30B-Thinking all had $\gamma = 0.81$ (Ge et al., 2026).[3] Much of the heterogeneity lies between studies: the 14 estimates of Ge et al. (2026) cluster around $\gamma = 0.82$, the direct choice estimates of Waisman (2026) are mostly much lower, and those of Hartley et al. (2025) are close to one. Without Ge et al., the remaining nine estimates pool to $\gamma = 0.65$ [0.33, 0.96]. At the study level, with four studies, the pooled $\gamma$ is 0.81 [0.37, 1.25], an interval that includes one. I therefore regard human-like probability weighting in language models as suggestive rather than established; non-parametric studies point in the same direction (Binz & Schulz, 2023; Hintze et al., 2025; Pozniak, 2026).

[3] For GPT-5.1, the 95% intervals that Ge et al. (2026) report in their Appendix C.5 ([1.28, 1.35] for curvature and [1.14, 1.22] for weighting) do not contain the point estimates of their main results table (0.87 and 0.81). They are identical to the intervals reported for Olmo-3-32B-Think, whose point estimates (1.33 and 1.16) they do contain, which suggests that the row for GPT-5.1 was misreported. I used the point estimates of the main table and imputed the standard errors for this model; excluding GPT-5.1 leaves the pooled curvature (1.01) and weighting (0.81) unchanged.

**Table 4. Probability weighting – nested meta-regression (γ; Tversky–Kahneman studies only).**

| | Model 1 | Model 2 | Model 3 | Model 4 | Model 5 |
|---|---|---|---|---|---|
| Intercept | 0.74 (0.16)*** | 0.79 (0.17)*** | — | 0.98 (0.18)*** | — |
| Qwen | 0.10 (0.23) | 0.03 (0.25) | — | -0.22 (0.26) | — |
| OLMo | 0.30 (0.28) | 0.16 (0.33) | — | -0.07 (0.33) | — |
| Claude | 0.15 (0.32) | 0.06 (0.34) | — | -0.05 (0.32) | — |
| DeepSeek | 0.08 (0.32) | -0.07 (0.37) | — | -0.30 (0.36) | — |
| Gemini | 0.11 (0.40) | -0.04 (0.45) | — | -0.27 (0.43) | — |
| Other | -0.43 (0.43) | 0.04 (0.73) | — | -0.77 (0.77) | — |
| Release year | | 0.10 (0.13) | — | -0.12 (0.16) | — |
| Stake size (log10) | | | — | 0.29 (0.15) | — |
| Persona conditioning | | | — | | — |
| Model fit (F) | F(6, 16) = 0.49 | F(7, 15) = 0.49 | — | F(8, 14) = 1.00 | — |
| $R^2$ | 0.00 | 0.00 | — | 0.00 | — |
| Estimates (k) | 23 | 23 | — | 23 | — |

Notes: Only studies fitted with the Tversky–Kahneman method are included. Moderators added in order: Model 1: model family, Model 2: release year, Model 3: gain–loss structure, not estimated because these gambles do not mix gains and losses, Model 4: stake size (log10, centred), and Model 5: persona conditioning. Model 5 (persona conditioning) could not be estimated, because no weighting estimate came from a persona prompt. No model is significant (all omnibus F tests $p > .47$). Coefficients from REML with the Knapp and Hartung adjustment; with four studies the table is exploratory. Other = RoBERTa. SEs in parentheses; *** $p < .001$.

### Study level results

At the study level, where each study contributes a single estimate (the mean of its estimates), the pooled loss aversion is $\lambda = 1.31$ [1.06, 1.56] across ten studies, close to the pool of the individual estimates (1.26): it is again well below the strongest human estimate ($t(9) = -4.94$, $p < .001$) and compatible with the intermediate one (1.31). Probability weighting under the Tversky–Kahneman method pools to $\gamma = 0.81$ [0.37, 1.25] across four studies, an interval too wide to separate it from either one or the human value. Value curvature under Tversky–Kahneman pools to 1.03 [0.94, 1.11] across six studies, again close to linear. In the multilevel model, the share of the heterogeneity that lies between studies, rather than between models within a study, was close to 0% for loss aversion, 15% for curvature and 45% for weighting: for loss aversion, the differences between models tested in the same study are as large as the differences between studies.

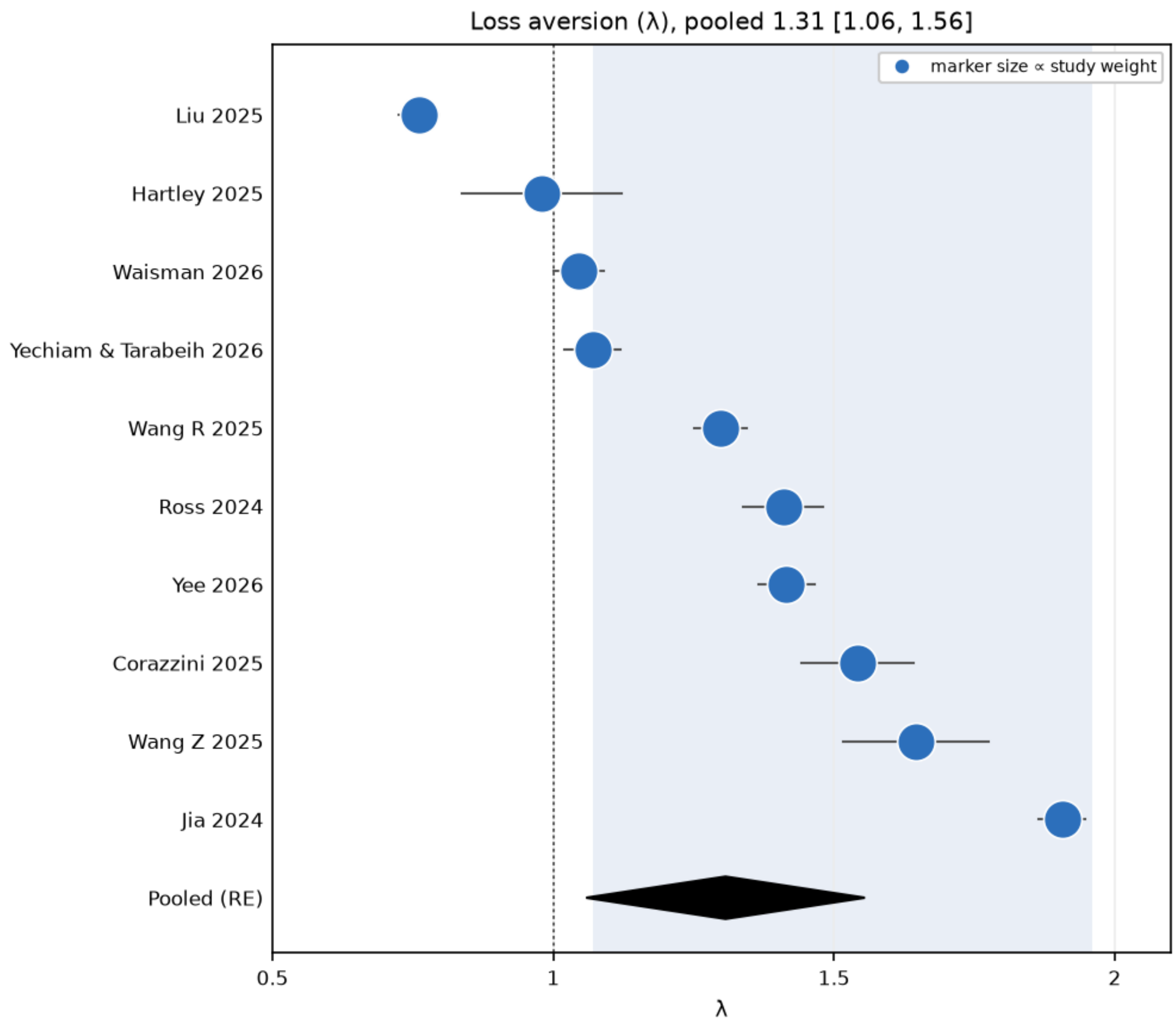

*Study-level forest plot – loss aversion (λ). Each row is one study's pooled estimate; the diamond is the random-effects pool. Values on the original λ scale; shaded band = human range (1.07–1.96).*

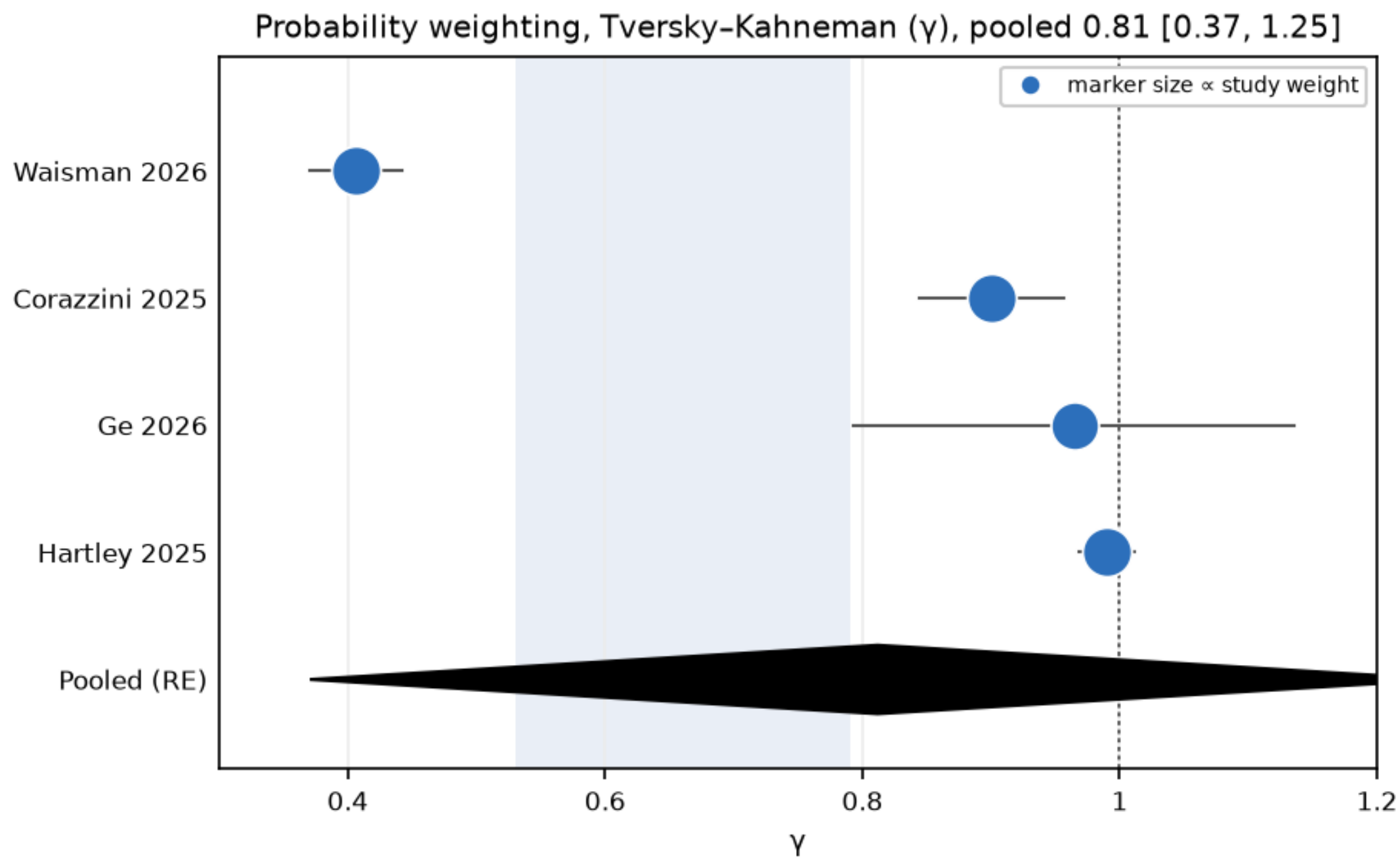

*Study-level forest plot – probability weighting, Tversky–Kahneman studies (γ). Shaded band = range of human estimates reported by Ruggeri et al. (2020), 0.53 to 0.79.*

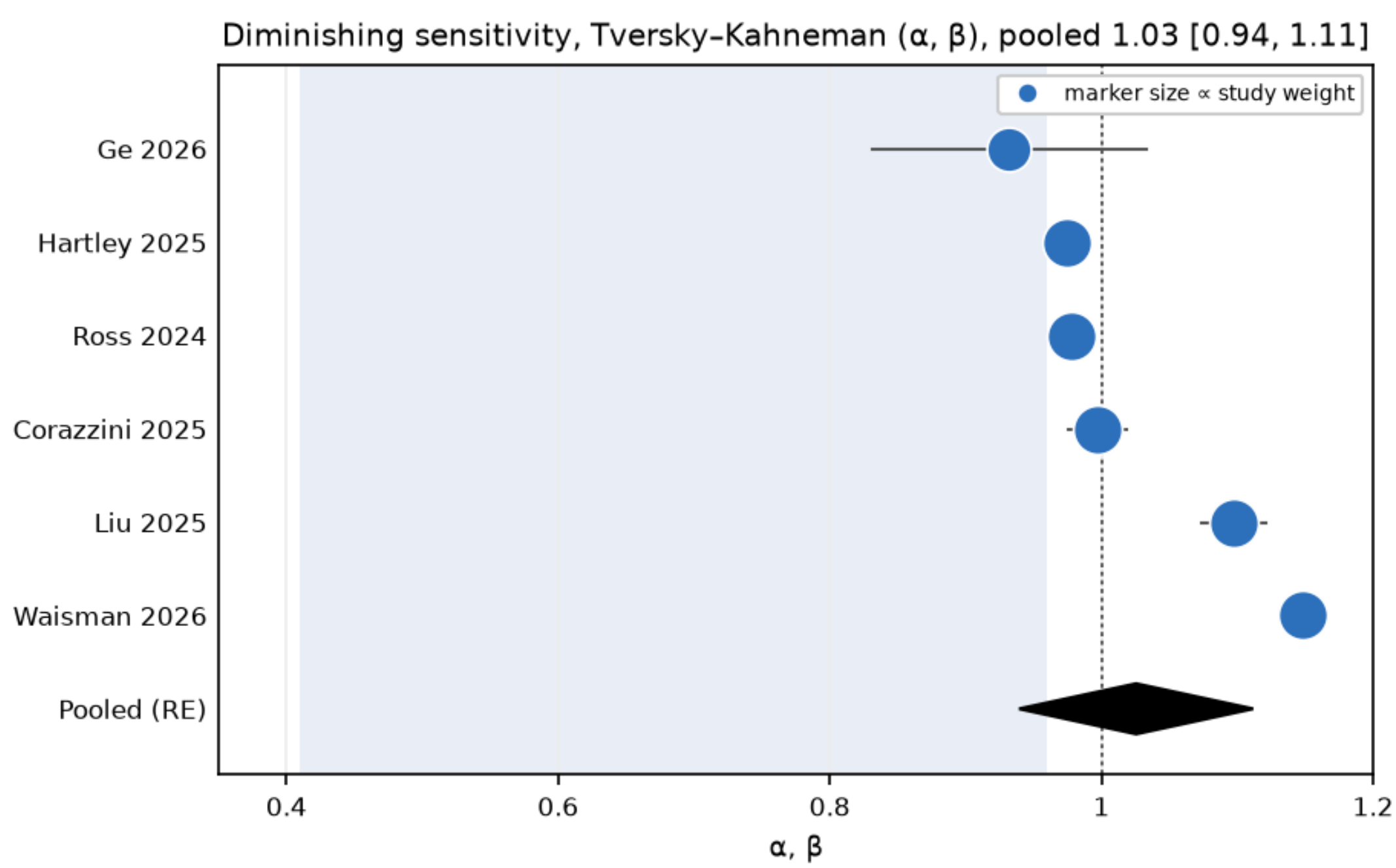

*Study-level forest plot – diminishing sensitivity, Tversky–Kahneman pool (α). Shaded band = range of human estimates reported by Ruggeri et al. (2020), 0.41 to 0.96.*

**Publication bias**

I assessed small study effects at the study level, the level at which publication and small study effects operate. Formal tests need about ten studies, so they were possible only for loss aversion. There, neither Egger's regression test (Egger et al., 1997) nor Begg's rank correlation test (Begg & Mazumdar, 1994) indicated funnel asymmetry (Egger $p = .58$; Begg $p = .37$), and the trim and fill method (Duval & Tweedie, 2000) imputed no missing studies. With six studies for curvature and four for weighting, the funnel plots are shown for completeness but cannot be tested, and the Tanaka pools come from only two studies. Even for loss aversion, ten studies give these tests little power, so the absence of evidence for publication bias should not be read as evidence of its absence.

Finally, I checked whether the six boundary estimates excluded in the Method section (four for loss aversion and two for probability weighting) affect the results, by adding them back. For loss aversion the pool moves to $\lambda = 1.42$ [1.00, 1.84], driven by two fits at the upper bound of the fitting routine (4.0 and 6.3), and loss aversion remains below the strongest human estimate; the excluded $\lambda$ values lie at both bounds (two at 0.01), so their exclusion does not push the pool in one direction. For probability weighting the two boundary estimates ($\gamma$ of 2.9 and 3.6, both from the Tanaka study and both at the upper bound) move the Tanaka pool from 0.84 [0.70, 0.98] to 1.32 [0.57, 2.07], so the inverse S result under the Tanaka method depends on excluding these fits. Diminishing sensitivity had no boundary estimates.

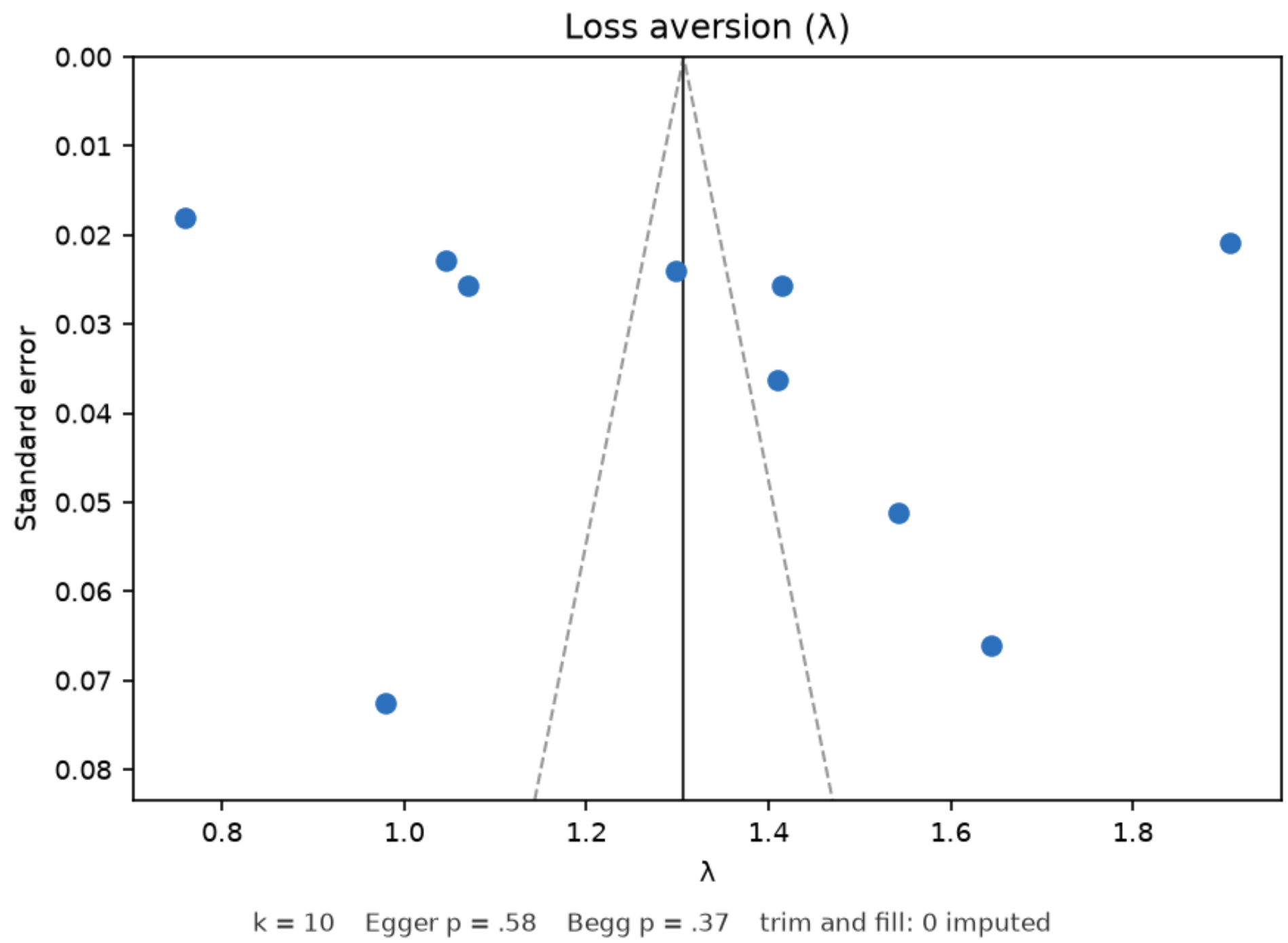

*Funnel plot for loss aversion (λ), by study (k = 10), original λ scale.*

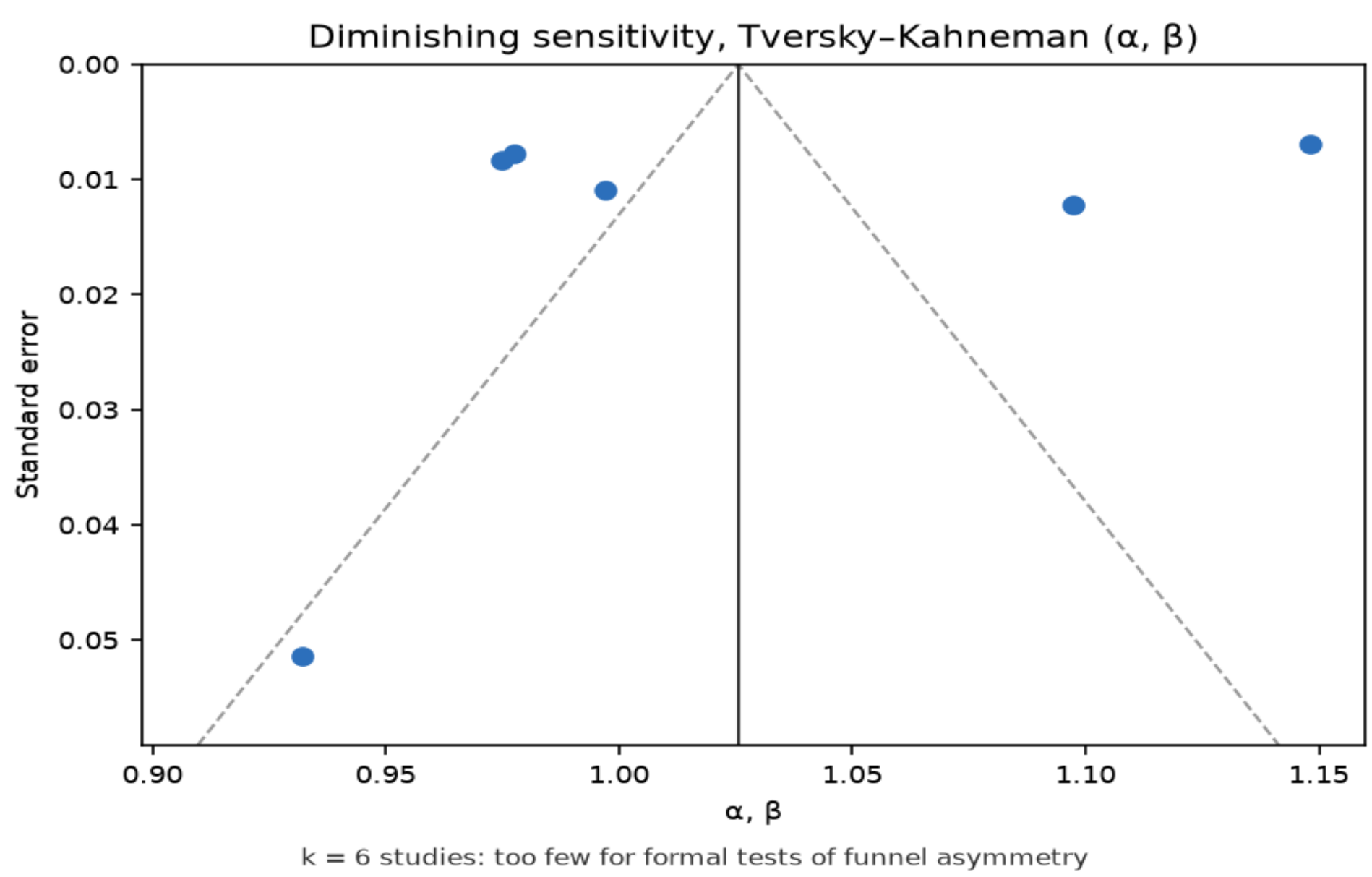

*Funnel plot for diminishing sensitivity – Tversky–Kahneman pool (α; k = 6). Too few studies for formal tests.*

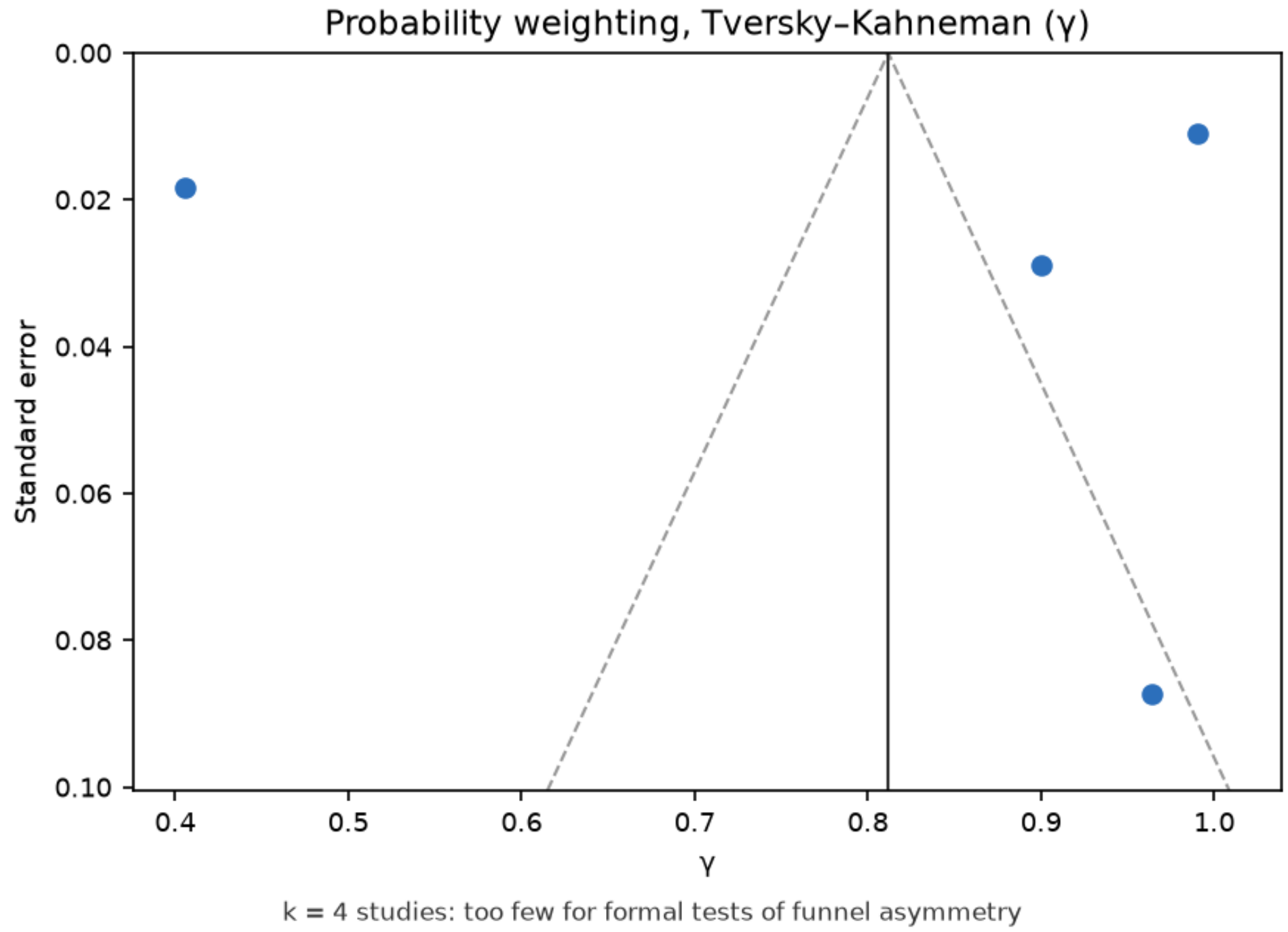

*Funnel plot for probability weighting, Tversky–Kahneman pool (γ; k = 4). Each study's standard error is the standard error of the mean of its estimates. Too few studies for formal tests.*

## Discussion

I pooled 102 fitted prospect theory parameters from 11 studies covering 32 language models. On average, the three components of prospect theory behave differently in language models than in people. Loss aversion is modest: the pooled coefficient ($\lambda = 1.26$) is far below the strongest human estimate (1.96) and compatible with the more conservative ones (1.07 and 1.31); compared with people tested in the same designs, it is compatible with the human value when gains and losses are symmetric and well below it when losses are smaller than gains. The value function is close to linear under the Tversky–Kahneman method (1.01) but more concave than in people under the Tanaka method (0.52), so diminishing sensitivity cannot yet be compared across fitting methods. Probability weighting is the most humanlike component, with the inverse S shape under both methods ($\gamma = 0.81$ and 0.84), although this conclusion rests on few studies and, under the Tanaka method, on the exclusion of two extreme fits.

The most important result, however, is not any of these averages but the size of the differences between models. The prediction interval for loss aversion runs from 0.20 to 2.31, and in the multilevel model nearly all of this variation lies between models tested in the same study. Model family explained about a quarter of it ($R^2 = .27$). Gemini models were the most loss averse, with estimates from 1.50 to 2.33 that fall inside or above the human range, and in the two studies that tested a Gemini model alongside GPT models it had the higher $\lambda$; the small open source models (Llama, Mistral, OLMo and RoBERTa) were mostly at or below loss neutrality (0.49 to 1.03). These family differences should be regarded as suggestive: they rest on few estimates per family, and the Gemini difference was only borderline once the dependence between estimates from the same study was taken into account. GPT models sat close to the overall average, although individual GPT models vary widely, from 0.57 (GPT-4.1) to 2.72 (GPT-3.5). When models differ this much, the pooled average describes no actual model: whether a language model resembles people is a question to be asked model by model, rather than about language models in general.

Why models differ is not yet known. They differ in their training data, in the fine tuning that shapes how they answer people, and in the safety and helpfulness objectives they are trained on, and any of these could shift the balance between gains and losses. The development of the models over time offers one clue, and it bears on the two hypotheses set out in the Introduction. Holding family constant, $\lambda$ fell by about 0.16 for every year of release, in line with the decline along the GPT lineage reported by Wang, Z. et al. (2025), from 2.72 in GPT-3.5 to 0.57 in GPT-4.1; this is the pattern the divergence hypothesis predicts. Probability weighting, by contrast, showed no trend, and there was no evidence that the newest reasoning models weighted probabilities differently from conversational models, although several of these fits reach the upper bound of the choice sensitivity parameter, that is, the models chose almost deterministically, and three reasoning models share the same estimate ($\gamma = 0.81$). The trend over time therefore cannot yet decide between the two accounts, but the components can. Where the accounts make different predictions for loss aversion, in designs with smaller losses than gains, the models were well below the human value and not reliably above loss neutrality ($\lambda = 1.46$ [0.96, 1.96]), as divergence predicts, and value curvature under the Tversky–Kahneman method was close to linear, again as divergence predicts. Probability weighting, in contrast, kept the human inverse S shape, as continuity predicts, although on few studies, and curvature under the Tanaka method was more concave than in people, which neither account predicts. Neither account therefore describes language models as a whole: they appear to keep some human regularities and lose others, and which ones they keep, and why, is the question the next studies should address

The study design moderators explained little. Persona conditioning did not explain the differences in loss aversion, and no moderator explained the differences in probability weighting. The gain–loss structure mattered less than in people, and the stakes gave only weak evidence of behaving as in people: $\lambda$ rose with the stakes once family and release year were held constant, but not on their own and not in the multilevel model. These effects rest on comparisons between a few studies whose designs are confounded with the models they tested, so they cannot fully separate properties of the models from properties of the tasks. What a model is shown also matters: persona and personality prompts shift its parameters (Ross et al., 2024; Liu et al., 2025; Hartley et al., 2025; Corazzini et al., 2025), and Yechiam and Tarabeih (2026) found that a model's loss aversion rises after it reads literature describing strong loss aversion (Supplementary Material S1). Part of what any study measures therefore reflects the prompt rather than a fixed preference.

These conclusions have clear limits. The corpus is small and several cells rest on a handful of estimates, so the confidence intervals are wide and the pooled values will shift as new studies appear. Heterogeneity is high wherever it can be measured. The family differences rest on a few estimates per family, so they should be confirmed by applying a single protocol to many models. Because most primary studies report a point estimate without a measure of uncertainty, the standard errors were mostly imputed; the imputed errors are small relative to the differences between models, so the random effects weights are nearly equal and the very high $I^2$ values partly reflect the imputation. The equal weight analyses (Table S2) and the multilevel analyses (Table S3) lead to the same averages, but the precision of the moderator effects should be read with caution. Finally, the difference between decisions from description and decisions from experience, which in people reverses the direction of rare event weighting (Barron & Erev, 2003; Hertwig & Erev, 2009; Wulff, Mergenthaler-Canseco & Hertwig, 2018), is so far covered by too few language model studies to pool, and is the most obvious place for new work.

In sum, language models are not uniformly more or less biased than people. The components of prospect theory come apart, and so do the model families. Probability weighting appears to be the most humanlike component; loss aversion is modest on average, but strong in some models and absent in others. For researchers who use language models in place of human participants, the practical lesson is that the choice of model matters as much as the design of the study: samples of responses from small open models and from Gemini models can lead to opposite conclusions about loss aversion. For those who use models to make or advise on real decisions, the same differences mean that a model's attitude toward losses should be measured rather than assumed.

**Acknowledgments**

I thank Eldad Yechiam for his guidance throughout the various stages of this research project and for helpful discussions and support.

**Appendix A. Parameterizations and standard-error imputation**

Value and weighting functions. Under the Tversky–Kahneman parameterization the value function is $v(x) = x^\alpha$ for gains and $-\lambda(-x)^\beta$ for losses, and probabilities are weighted by $w(p) = p^\gamma / (p^\gamma + (1-p)^\gamma)^{1/\gamma}$. Under the Tanaka–Camerer–Nguyen parameterization a single curvature parameter governs the value function. It is written in two ways in the literature: Tanaka, Camerer and Nguyen (2010) and Wang, R. et al. (2025) use the power exponent itself, $v(x) = x^\sigma$ for gains and $-\lambda(-x)^\sigma$ for losses, so that a lower σ means a more concave function, whereas Jia et al. (2024) write $v(x) = x^{(1-\sigma)}$. I converted the estimates of Jia et al. to the exponent $(1 - \sigma)$ before pooling, so all Tanaka curvature values in this paper are exponents, comparable with the human value of 0.67. In this parameterization probabilities are weighted by the Prelec function $w(p) = \exp(-(-\ln p)^\gamma)$. λ is the loss-aversion coefficient in both. Ge et al. (2026) fit a related four parameter model with separate choice sensitivity parameters for gains and losses; its curvature parameter, which they call σ, plays the role of α, and I pooled it with the Tversky–Kahneman curvature estimates. The Tversky–Kahneman γ and the Prelec γ both capture the inverse S shape (lower values mean a stronger distortion), but they are not on exactly the same scale, so γ values are compared only within each parameterization.

Standard-error imputation. Most studies report a point estimate without an interval. For those estimates I imputed a standard error with a rule adapted from Furukawa et al. (2006): $SE = SD / \sqrt{n}$, where SD is the standard deviation of the estimates in the pooled set and n is the number of runs behind the estimate. When a study reported a 95% confidence interval (Hartley et al., 2025; Wang, R. et al., 2025; Ge et al., 2026, Appendix C.5), the standard error was the width of the interval divided by 3.92; for Jia et al. (2024), who report the standard deviation of each parameter across their 300 runs, it was that standard deviation divided by $\sqrt{300}$. One interval in Ge et al. (2026), for GPT-5.1, does not contain the point estimate reported in their main table, so for that model the standard error was imputed; the results are unchanged without GPT-5.1 (curvature 1.01, weighting 0.81). All three parameters were pooled on their natural scale, with REML and the Knapp and Hartung adjustment.

## Supplementary Material

### S1. Sensitivity of loss aversion to the described literature

One more result bears on how to read any of these parameters. Yechiam and Tarabeih (2026), a study I coauthored that is included in the meta-analysis, show that the loss aversion a model displays depends on the literature it is given. Given Brown et al. (2024), which reports strong loss aversion, ChatGPT and Qwen then behaved in a loss-averse way; given Yechiam and Zeif (2025), which reports weak or absent loss aversion for symmetric lotteries, the same models returned λ between 1.0 and 1.1, close to their unprompted baselines of 1.05 and 1.09. The stronger the bias described in the input, the stronger the bias in the model's own later choices.

This has a knock-on implication for the present work: a meta-analysis of loss aversion in language models is partly a meta-analysis of the literature those models have read. It also hints at why the three components come apart. A parameter that shifts when you describe the relevant literature is better thought of as a response drawn out by the prompt than as a fixed preference, and the prompting conditions behind most published estimates are neither standardized nor fully reported.

## Supplementary tables

**Table S1. Random-effects meta-analysis of prospect-theory parameters against human benchmarks.**

| Parameter and subset | k | Random-effects pooled (95% CI; PI = prediction interval) | Human | Test vs human; heterogeneity |
|---|---|---|---|---|
| Loss aversion λ | 28 | 1.26 (1.06–1.46); PI 0.20–2.31 | 1.96 (Brown) | t = −5.78, p < .001; τ = 0.51; Q(27) = 3,875, I² = 99% |
| equal-weight, by model (robustness) | 19 | 1.26 (1.02–1.49) | 1.96 | t(18) = −6.26, p < .001, d = −1.44 |
| Diminishing sensitivity, T&K | 33 | 1.01 (0.95–1.06); PI 0.69–1.33 | 0.67 (Imai); 0.88 (T&K) | t = 12.41 and 4.72, both p < .001; τ = 0.15; Q(32) = 2,473, I² = 99% |
| Diminishing sensitivity, Tanaka | 10 | 0.52 (0.46–0.58) | 0.67 (Tanaka) | t = −5.43, p < .001 (exponent; lower = more concave); Q(9) = 6,813, I² = 100% |
| Weighting, T&K | 23 | 0.81 (0.65–0.98); PI 0.06–1.56 | 0.68 (Imai); 0.65 (T&K) | t = 1.64, p = .12, and t = 2.03, p = .055; τ = 0.35; Q(22) = 2,029, I² = 99% |
| T&K, excluding Ge et al. (2026) | 9 | 0.65 (0.33–0.96) | 0.68 (Imai) | t = −0.25, p = .81; Q(8) = 1,319, $I^2$ = 99% |
| Weighting, Tanaka | 8 | 0.84 (0.70–0.98) | 0.685 (Tanaka) | t = 2.57, p = .037; Q(7) = 2,224, $I^2$ = 100% |

Note. Each row is a random effects meta-analysis (REML with the Knapp and Hartung adjustment) over all estimates in that subset; k is the number of estimates. Standard errors are reported intervals where available and are otherwise imputed (Appendix A). Tests against the human benchmarks are t tests that include the uncertainty of the benchmark where the source reports an interval (Brown et al., 2024; Imai et al., 2025, for γ). Heterogeneity is τ, Cochran's Q and $I^2$. Because most standard errors are imputed, Q and $I^2$ mainly reflect how these errors were constructed and are not interpretable as measures of heterogeneity; τ and the prediction interval are. The equal weight, by model loss aversion row, and Table S2, give the corresponding equal weight results as a robustness check; the "excluding Ge" row drops the 14 estimates of Ge et al. (2026). Human benchmarks: Brown et al. (2024) for λ; Imai et al. (2025) and Tversky and Kahneman (1992) for the Tversky–Kahneman parameters; Tanaka et al. (2010) for the Tanaka parameters, with Tanaka curvature expressed as the exponent of the power value function (lower = more concave).

## Table S2. Equal weight means as a function of the analytic unit.

| Parameter | By estimate | By model | By study |
|---|---|---|---|
| Loss aversion λ | 1.25 (1.06–1.45), k = 28 | 1.26 (1.02–1.49), k = 19 | 1.31 (1.06–1.56), k = 10 |
| Weighting γ, T&K | 0.84 (0.66–1.02), k = 23 | 0.91 (0.74–1.08), k = 19 | 0.82 (0.38–1.25), k = 4 |

Note. Equal weight means with t based 95% confidence intervals. Weighting is pooled within the Tversky–Kahneman studies only, in line with the pooling rule of the main analysis.

## Table S3. Loss aversion: nested meta-regressions with estimates nested in studies (multilevel model, cluster robust standard errors).

| | Model 1 | Model 2 | Model 3 | Model 4 | Model 5 |
|---|---|---|---|---|---|
| Intercept | 1.22 (0.13)*** | 1.22 (0.13)*** | 1.16 (0.13)** | 1.10 (0.14)* | 1.08 (0.19) |
| Claude | 0.00 (0.18) | 0.12 (0.14) | 0.17 (0.13) | 0.11 (0.22) | 0.15 (0.27) |
| Gemini | 0.69 (0.29) | 0.81 (0.20)* | 0.76 (0.23)* | 0.85 (0.33) | 0.86 (0.50) |
| Qwen | 0.22 (0.19) | 0.30 (0.19) | 0.29 (0.18) | 0.35 (0.15) | 0.35 (0.19) |
| Other families | −0.49 (0.16)* | −0.62 (0.29) | −0.60 (0.33) | −0.63 (0.30) | −0.89 (0.71) |
| Release year | | −0.18 (0.19) | −0.18 (0.22) | −0.22 (0.23) | −0.25 (0.33) |
| Losses smaller than gains (L < G) | | | 0.18 (0.16) | 0.27 (0.16) | 0.30 (0.26) |
| Stake size (log10) | | | | 0.35 (0.22) | 0.31 (0.26) |
| Persona conditioning | | | | | 0.49 (0.81) |
| Model fit (F) | F(4, 5) = 15.08** | F(5, 4) = 6.77* | F(6, 3) = 3.32 | F(7, 2) = 4.71 | F(8, 1) = 104.78 |
| $R^2$ | 0.27 | 0.37 | 0.36 | 0.49 | 0.50 |
| Estimates (k) / studies (m) | 28 / 10 | 28 / 10 | 28 / 10 | 28 / 10 | 28 / 10 |

Note. Multilevel model (REML) with estimates nested in the ten studies; cluster robust standard errors with the small sample factor $m/(m - p)$ and t tests with $m - p$ degrees of freedom, so Model 5 ($p = 9$ coefficients) has a single degree of freedom; Models 3 to 5 are therefore descriptive. Moderators as in Table 2. Leaving out one study at a time (not tabulated) kept the pooled λ between 1.21 and 1.30 and the Gemini coefficient between +0.50 and +0.90; the release year coefficient was not significant in any of these analyses, and without Wang, Z. et al. (2025) it fell to −0.08. With study fixed effects the family coefficients were Claude +0.26, Gemini +0.67, Qwen +0.47 and other families −0.34 (all $p > .09$). Without RoBERTa the pooled λ was 1.27 [1.07, 1.48]. In the multilevel model, model family was significant (Model 1: $F(4, 5) = 15.08$, $p = .005$, $R^2 = .27$); the Gemini difference was significant only once release year was held constant (Model 2: +0.81, $p < .05$), and the small open source models were lower in Model 1 (−0.49, $p = .03$). Release year had a similar coefficient to that in Table 2 (−0.18 in Model 2) but a standard error nearly three times larger (0.19, $p = .40$). In the two studies that tested a Gemini model alongside GPT models (Jia et al., 2024; Yee & Koh, 2026), the Gemini model had a higher λ than the GPT models, and in the third (Wang, R. et al., 2025) it was second only to Qwen2.5-14B.